\pdfoutput=1

\documentclass[11pt]{article}

\usepackage[]{acl}

\usepackage{times}
\usepackage{latexsym}

\usepackage[T1]{fontenc}

\usepackage[utf8]{inputenc}

\usepackage{microtype}

\usepackage{inconsolata}

\usepackage{times}
\usepackage{soul}
\usepackage{url}
\usepackage{graphicx}
\usepackage{amsmath}
\usepackage{amsthm}
\usepackage{booktabs}
\usepackage{algorithm}
\usepackage{algorithmic}
\usepackage{inconsolata}
\usepackage{import}
\usepackage{microtype}
\usepackage{layout}
\usepackage{tabularx, makecell}
\usepackage{booktabs}
\usepackage{mathrsfs}
\usepackage{amssymb} 
\usepackage{url}
\usepackage{xspace,paralist}
\usepackage{times,latexsym}
\usepackage{amsmath}
\usepackage{appendix}
\usepackage{comment} 
\usepackage{enumitem}
\usepackage{makecell}
\usepackage{multirow}
\usepackage{xcolor}
\usepackage{cleveref}
\usepackage{tcolorbox}
\usepackage{adjustbox}
\usepackage{relsize}
\usepackage{todonotes}
\usepackage{natbib}

\newcommand{\tabref}[2][]{Table#1~\ref{#2}\xspace}
\newcommand{\figref}[1]{Figure~\ref{#1}\xspace}
\newcommand{\secref}[1]{Section~\ref{#1}\xspace}

\newcommand{\model}[1]{\text{#1}\xspace}

\newcommand{\chatgpt}{\model{ChatGPT}}
\newcommand{\gptfour}{\model{GPT-4}}

\newcommand{\pplai}{\model{PerplexityAI}}

\newcommand{\checker}[1]{\textit{#1}\xspace}
\newcommand{\snowball}{\checker{Snowball}}
\newcommand{\freshqa}{\checker{FreshQA}}
\newcommand{\selfaware}{\checker{SelfAware}}

\newcommand{\factool}{\checker{FacTool}}
\newcommand{\felmwk}{\checker{FELM-WK}}
\newcommand{\factcheckgpt}{\checker{Factcheck-GPT}}
\newcommand{\halueval}{\checker{HaluEval}}
\newcommand{\factor}{\checker{FACTOR}}
\newcommand{\rarr}{\checker{RARR}}
\newcommand{\factscore}{\checker{FactScore}}
\newcommand{\cove}{\checker{CoVe}}

\usepackage{amssymb}
\usepackage{pifont}
\newcommand{\cmark}{\ding{51}}%
\newcommand{\xmark}{\ding{55}}%

\usepackage[switch]{lineno}

\title{Factuality of Large Language Models: A Survey}

\author{
    Yuxia Wang$^{1}$,\quad Minghan Wang$^{2}$,\quad Muhammad Arslan Manzoor$^{1}$,\quad Fei Liu$^{3}$, \\
    \textbf{Georgi Georgiev}$^{4}$,\quad \textbf{Rocktim Jyoti Das}$^{1}$,\quad \textbf{Preslav Nakov}$^{1}$ \\
    \vspace{0.8mm}
     $^{1}$MBZUAI, \quad $^{2}$Monash University, \quad $^{3}$Google, \quad $^{4}$Sofia University \\
     \texttt{\{yuxia.wang, preslav.nakov\}}@mbzuai.ac.ae
}

\begin{document}
\maketitle
\begin{abstract}
Large language models (LLMs), especially when instruction-tuned for chat, have become part of our daily lives, freeing people from the process of \emph{searching}, \emph{extracting}, and \emph{integrating} information from multiple sources by offering a straightforward answer to a variety of questions in a single place.
Unfortunately, in many cases, LLM responses are factually incorrect, which limits their applicability in real-world scenarios.
As a result, research on evaluating and improving the factuality of LLMs has attracted a lot of attention recently.
In this survey, we critically analyze existing work
with the aim to identify the major challenges and their associated causes, pointing out to potential solutions for improving the factuality of LLMs, and analyzing the obstacles to automated factuality evaluation for open-ended text generation. We further offer an outlook on where future research should go.

\end{abstract}

\section{Introduction}
Large language models (LLMs) have become an integral part of our daily lives. When instruction-tuned for chat, they have enabled digital assistants that can free people from the need to \emph{search}, \emph{extract}, and \emph{integrate} information from multiple sources by offering straightforward answers in a single chat.
While people naturally expect LLMs to always present reliable information that is consistent with real-world knowledge, LLMs tend to fabricate ungrounded statements, resulting in misinformation~\cite{tonmoy2024survey}, which limits their utility.
Thus, assessing and improving the factuality of the text generated by LLMs has become an emerging and crucial research area, aiming to identify potential errors and to advance the development of more reliable LLMs~\cite{chen2023felm}. 

To this end, researchers have collected multiple datasets, introduced a variety of measures to evaluate the factuality of LLMs, and proposed numerous strategies leveraging external knowledge through retrieval, self-reflection, and early refinement in model generation to mitigate factual errors~\cite{tonmoy2024survey}. Numerous surveys~\citep{tonmoy2024survey, Huang2023SurveyHalLLM, Wang2023SurveyOF} have explored factuality or hallucinations in large language models across various modalities. While they either lack in-depth discussion or are too specific to grasp the fundamental challenges, promising solutions in factuality evaluation and enhancement, and some ambiguous concepts in LLM factuality. 
We summarized these surveys in Table~\ref{tab:surveys}.

Our survey aims to bridge this gap by providing an in-depth analysis of LLM factuality, with an emphasis on recent studies to reflect the rapidly evolving nature of the field. We offer a comprehensive overview of different categorizations, evaluation methods, and mitigation techniques for LLM factuality in both language and vision modalities. Additionally, we explore a novel research avenue that seeks to improve LLM calibration. This includes making models aware of their knowledge limitations and enhancing the reliability of their output confidence.

\begin{table*}[t!]
  \centering
  \small
  \resizebox{\textwidth}{!}{
  \begin{tabular}{l cc ccc p{11cm}}
    \toprule
    \textbf{Survey} & \textbf{Date} & \textbf{Pages} &  \textbf{Eval} & \textbf{Improve} & \textbf{Multimodal} & \textbf{Contributions and limitations} \\
    \midrule 
    Our work & 15-June-2024 & 9 & \cmark & \cmark & \cmark & Discusses ambiguous concepts in LLM factuality, compares and analyzes evaluation and enhancement approaches from academic and practical perspectives, outlining major challenges and promising avenues to explore. \\
    \cite{tonmoy2024survey} & 08-Jan-2024 & 19 & \xmark & \cmark & \cmark & Summarizes recent work in terms of mitigating LLM hallucinations, but \textbf{lacks comparison} between different approaches and \textbf{discussions} to identify open questions and challenges. \\
    \cite{gao2023RAG} & 18-Dec-2023 & 26 & \xmark & \cmark & \xmark & Summarizes three \textbf{RAG paradigms}: na\"{i}ve, advanced, and modular RAG, with key elements and evaluation methods for the three major components in RAG (retriever, generator, and augmentation). \\
    \cite{huang2023survey} & 09-Nov-2023 & 49 & \cmark & \cmark & \xmark & Analyzes the reasons for hallucinations, and presents a comprehensive overview of hallucination detection methods, benchmarks, and approaches to mitigate hallucinations. \\
    \cite{Wang2023SurveyOF} & 18-Oct-2023 & 44 & \cmark & \cmark & \xmark & Detailed literature review of factuality improvement and enhancement methods covering both retrieval augmentation and non-retrieval augmentation, missing discussion of major bottleneck issues in LLM factuality and promising directions to investigate.  \\
    \cite{rawte2023hallucinationsurvey} & 18-Sept-2023 & 11 & \xmark & \xmark & \cmark & Extensively elucidates the problem of hallucination across all major modalities of foundation models, including text (general, multilingual, domain-specific LLMs), image, video, and audio. However, inadequate coverage of approaches, in-depth categorization and comparison between methods. \\
    \cite{zhang2023siren} & 03-Sept-2023 & 32 & \cmark & \cmark & \xmark & Organized by different training stages of LLMs, discusses potential sources of LLM hallucinations and in-depth review of recent work on addressing the problem. \\
   \cite{guo-etal-2022-survey} & Feb-2022 & 29 & \cmark & \xmark & \xmark  & Focused on the automated fact-checking pipeline \\
    \bottomrule
  \end{tabular}
  }
  \caption{Comparison of different surveys on the factuality of LLMs. Eval: Evaluation; Improve: Improvement.}
  \label{tab:surveys}
\end{table*}

\section{Background}
\label{sec:background}
Hallucination and factuality, while conceptually distinct, often occur in similar contexts and are sometimes used interchangeably,
rendering them intricately intertwined, posing a challenge in discerning their distinct boundaries, and causing a considerable amount of misconception.
In this section, we seek to disambiguate and refine our understanding of these two closely aligned concepts, thereby preventing misinterpretation and reducing potential confusion.
Additionally, we further include two closely-related axes: relevance and trustworthiness for LLM evaluation to illustrate their nuance in relation to factuality.

\paragraph{Hallucination vs. Factuality}
The concept of hallucination in the context of traditional natural language generation tasks is typically referred to as the phenomenon in which the generated content appears nonsensical or unfaithful to the provided source content~\cite{ji2023-nlg-survey}. 
One concrete example is made-up information in an abstractive summary with additional insights beyond the scope of the original source document.

In the age of LLMs, the term hallucination has been reimagined, encompassing any deviation from factual reality or the inclusion of fabricated elements within generated texts~\cite{tonmoy2024survey,rawte2023hallucinationsurvey}.
\cite{zhang2023siren} define hallucination as the characteristic of LLMs to generate content that diverges from the user input, contradicts previously generated context, or mis-aligns with established world knowledge.
\cite{huang2023survey} merge the input- and context-conflicting types of hallucinations and further take logical inconsistency into account to form \textit{faithfulness} hallucination.
Another category is \textit{factuality} hallucination, referring to the discrepancy between generated content and verifiable real-world facts, manifesting as (1) factual inconsistency and (2) factual fabrication.

Factuality, on the other hand, is concerned with a model's ability to learn, acquire, and utilize factual knowledge.
\cite{Wang2023SurveyOF} characterize factuality issues as the probability of LLMs producing content inconsistent with established facts.
It is important to note that hallucination content may not always involve factual missteps.
Though a piece of generated text may exhibit divergence from the initial prompt’s specifics, it falls into hallucinations, not necessarily a factual issue if the content is accurate.

It is crucial to distinguish between factual errors and instances of hallucination.
The former involves inaccurate information whereas the latter can present unanticipated and yet factually substantiated content~\cite{Wang2023SurveyOF}.

\textbf{Summary:}
Factuality is the ability of LLMs to generate content consistent with factual information and world knowledge. Although both hallucinations and factuality may impact the credibility of LLMs in the context of content generation, they present distinct challenges. Hallucinations occur when LLMs produce baseless or untruthful content, not grounded in the given source. In contrast, factuality errors arise when the model fails to accurately learn and utilize factual knowledge. It is possible for a model to be factually correct yet still produce hallucinations by generating content that is either off-topic or more detailed than what is requested.


\paragraph{Trustworthiness/Reliability vs. Factuality} 
In the context of LLMs, factuality~\cite{Wang2023SurveyOF} refers to a model's capability of generating contents of factual information, grouneded in reliable sources (e.g., dictionaries, Wikipedia or textbooks), with commonsense, world and domain-specific knowledge taken into account.
In contrast, ``trustworthiness'' \cite{Sun2024TrustLLMTI} extends beyond mere factual accuracy and is measured on eight dimensions: truthfulness, safety, fairness, robustness, privacy, ethics, transparency, and accountability.


\begin{table*}[t!]
  \centering
  \small
  \resizebox{\textwidth}{!}{
  \begin{tabular}{cllrr l c}
    \toprule
    \textbf{Type} & \textbf{Dataset} & \textbf{Topic} & \textbf{Size} & \textbf{ER\%} & \textbf{Evaluation and Metrics} used in Original Paper & \textbf{Freq} \\
    \midrule 
    \multirow{7}{*}{I} 
    & \factscore-Bio~\cite{min2023factscore} & Biography & 549 & 42.6 & Human annotation and automated fact-checkers & 4 \\
    & \factcheckgpt~\cite{wang2023factcheckgpt} & Open-ended questions & 94 & 64.9 & Human annotation & 1 \\
    & \factool-QA~\cite{chern2023factool} & Knowledge-based QA & 50 & 54.0 & Human annotation and automated fact-checkers & 2 \\
    & \felmwk~\cite{chen2023felm} & Knowledge-based QA & 184 & 46.2 & Human annotation, Accuracy and F1 score & 1 \\
    & \halueval~\cite{li2023halueval} & Open-ended questions & 5000 & 12.3 & Human annotation, AUROC + LLM judge + PARENT & 3 \\
    & \freshqa~\cite{vu2023freshllms} & Open-ended questions & 499 & 68.0 & Human annotation & 2 \\
    & \selfaware~\cite{yin2023selfaware} & Open-ended questions & 3369  & -- & Evaluate the LLM awareness of unknown by F1-score & 1\\
    \midrule
    II & \snowball~\cite{zhang2023snowball} & Yes/No question & 1500 & 9.4 & Exact match + Accuracy/F1-score & 1 \\
    \midrule 
    \multirow{2}{*}{III} 
    & \textit{Wiki-category List}~\cite{dhuliawala2023chain} & Name some \textit{[Mexican films]} & 55 & -- & Precision/recall@5 & 1\\
    & \textit{Multispan QA}~\cite{dhuliawala2023chain} & Short-term Answer & 428 & -- & Exact match + F1 score & 1\\
    \midrule 
    \multirow{4}{*}{IV} 
    & TruthfulQA~\cite{lin2022truthfulqa} & False belief or misconception & 817 & -- & Accuracy & 5 \\
    & HotpotQA~\cite{yang2018hotpotqa} & Multi-step reasoning& 113k & -- & Exact match + F1 score & 11 \\
    & StrategyQA~\cite{geva2021strategyqa} & Multi-step reasoning& 2780 & -- & Recall@10 & 3 \\
    & MMLU~\cite{hendrycks2021mmlu} & Knowledge & 15700 & -- & Accuracy & 4 \\
    \bottomrule
  \end{tabular}
  }
  \caption{\textbf{Four types of datasets used to evaluate LLM factuality}. I: open-ended generation; II: Yes/No answer; III: short-term or list of entities answer; IV: A, B, C, D multiple Choice QA. Labeled datasets under type I are mostly generated by \chatgpt, and \factscore-Bio (\chatgpt, \textit{InstGPT} and \pplai). ER: Human-annotated Error Rate. Freq: usage frequency as evaluation set in our first 50 references.}
  \label{tab:datasets-info}
\end{table*}

\section{Evaluating Factuality}
\label{sec:evaluation}
Evaluating LLM factuality on open-ended generations presents a non-trivial challenge, discerning the degree to which a generated textual statement aligns with objective reality.  
Studies employ various benchmarks, evaluation strategies and metrics to achieve this goal. 

\subsection{Datasets and Metrics}
While \cite{zhang2023siren} outlined tasks and measures for hallucination evaluation, there is no comparative analysis of existing datasets to assess various aspects in regards to model factuality (e.g., knowledge grounding, fast-changing facts, snowballing hallucinations, robustness to false premises, and uncertainty awareness).
We categorize the datasets in the format of discrimination or generation, and highlights the challenges in automatic evaluation for long-form open-ended generations.

Current benchmarks largely assess the factuality in LLMs based on two capabilities: proficiency in distinguishing factual accuracy in a context and ability to generate factually sound content.

The former typically comes in the form of a multi-choice question, with the expected response being a label of one of A, B, C, and D.
For instance, HotpotQA, StrategyQA, MMLU.
This form of evaluation has been widely used to measure the general knowledge proficiency and factual accuracy of LLMs, largely thanks to its automation-friendly nature.
Under this evaluation formulation, model responses are easily parsed and compared with gold standard labels, enabling the calculation of accuracy or F1 scores against established benchmarks.

Precisely assessing the factuality of free-form LLM outputs remains a significant challenge due to the inherent limitations of automatic methods in the face of open-ended generation and the absence of definitive gold standard responses within an expansive output space.
To make automatic evaluation feasible, many studies constrain the generation space to (1) Yes/No; (2) short-form phrase; and (3) a list of entities through controlling the categories of questions and generation length.

Perhaps the most demanding, yet inherently realistic scenario is free-form long text generation, such as biography generation.
For this, the most commonly used and reliable methods rely on human experts following specific guidelines, and automatic fact-checkers based on retrieved information, such as FactScore, Factool and Factcheck-GPT, to facilitate efficient and consistent evaluation.

These automatic fact-checkers generally first decompose a document into a set of atomic claims, and then verify one by one whether the claim is true or false based on the retrieved evidence, either from offline Wikipedia or online Web pages.
The percentage of true claims over all statements in a document is used to reflect the factual status of a response (refer to FactScore). The averaged Factscore over a dataset is in turn used to assess a model's factuality accuracy.
However, there is no guarantee that automatic fact-checkers are 100\% accurate in their verification process.
\cite{wang2023factcheckgpt} show that even the state-of-the-art verifier, equipped with GPT-4 and supporting evidence retrieved with Google search, can only achieve an F1 score of $0.63$ in identifying false claims and F1=0.53 using \pplai  (compared with human-annotated labels for claims: true or false). 

\textbf{Summary:} We categorize datasets that evaluate LLM factuality into four types, depending on the answer space and the difficulty degree on which accurate automatic quantification can be performed (see \tabref{tab:datasets-info}).
They are: 
(I) open-domain, free-form, long-term responses (FactScore: the percentage of the correct claims verified by human or automated fact-checker);  
(II) Yes/No answer w/wt explanation (extract Yes/No, metrics for binary classification);
(III) short-form answer (Exact match the answer with gold labels and calculate accuracy) or the listing answer (recall@K); and
(IV) multi-choice QA (metrics for multi-class classification).

\subsection{Other Metrics}
\label{sec:othermetrics}
In addition to evaluating the methods discussed above, 
\cite{lee2022factuality} quantified the hallucinations using two metrics, both requiring document-level ground-truth:
(1) \emph{hallucinated named entities error} measures the percentage of named entities in the generations that do not appear in the ground-truth document;
(2) \emph{entailment ratio} evaluates the number of generations that can be entailed by the ground-truth reference, over all generations. 

\cite{Rawte2023HalLLMRemed} defined the hallucination vulnerability index (HVI), which takes a spectrum of factors into account, to evaluate and rank LLMs.

Some factuality measurement tasks, such as claim extraction and evidence retrieval are non-trivial to automate. 
\cite{Rawte2023HalLLMRemed} curated publicly available LLM hallucination mitigation benchmark, where LLM generations are scored by humans when automated external knowledge retrieval fails to resolve a claim clearly. 
While widely used for factuality evaluation, this hybrid approach 
may suffer from human annotation bias.

\section{Improving Factuality }
\label{sec:enhancement}
Improving the factuality of an LLM often requires updating its internal knowledge, editing fake, outdated and biased elements, thereby making its output reflect a revised collection of facts, maximizing the probability of $P(\textrm{truth}|\textrm{prompt})$.
One option is to adopt gradient-based methods to update model parameters to encourage desired model output. This includes pre-training, supervised fine-tuning and RLXF.
We can also explore injecting a new fact into LLMs or overwriting the false knowledge stored in LLM memory by in-context learning (ICL).
When models store factually correct knowledge but produce errors, they can in some cases rectify them through self-reasoning, reflection, and multi-agent debates.

We discuss these methods throughout the lifecycle of an LLM, ranging from pre-training, to inference, to post-processing.
Another important element is retrieval augmentation, which enhances the generation capabilities of LLMs by anchoring them in external knowledge that may not be stored or contradict the information in LLM parametric memory.
It can be incorporated at various stages throughout model training and the subsequent inference process~\cite{gao2023RAG}, and is therefore not discussed individually.
\vspace{-1mm}
\subsection{Pre-training}
\vspace{-1mm}
LLMs store a vast amount of world knowledge in their parameters through the process of pre-training.
The quality of the pre-training data plays a crucial role and misinformation could potentially cause LLMs to generate false responses, motivating the utilization of high-quality textual corpora.
However, the prohibitively massive amount of pre-training data, typically consisting of trillions of tokens, renders manual filtering and editing impractically laborious.
To this end, automated filtering methods have been proposed.
For instance, \cite{brown2023gpt3} introduce a method to only focus on a small portion of the CommonCrawl dataset that exhibits similarity to high-quality reference corpora.
\cite{touvron2023llama2} propose to enhance factual robustness of mixed corpora by up-sampling documents from the most reliable sources, thereby amplifying knowledge accuracy and mitigating hallucinations.
During the pre-training phase of \textbf{phi-1.5}, \cite{li2023phi} synthesize ``textbook-like'' data, consists of and rich in high-quality commonsense reasoning and world knowledge.
While careful corpus curation remains the cornerstone of pre-training for enhanced factuality, the task becomes increasingly challenging with the expansion of dataset scale and the growing demand for linguistic diversity. 
It is therefore crucial to develop novel strategies that guarantee the consistency of factual knowledge across diverse cultural landscapes.

\cite{Borgeaud2021ImprovingLM} propose RETRO, a retrieval augmented pre-training approach. An auto-regressive LLM is trained from scratch with a retrieval module that is practically scalable to large-scale pre-training by retrieving billions of tokens. RETRO shows better accuracy and is less prone to hallucinate compared to GPT~\cite{Wang2023ShallWP}. While limitations lie in that RETRO performance could be compromised if the retrieval database contains inaccurate, biased or outdated information. $\sim$25$\%$ additional computation is required for the pre-training of LLMs with retrieval. 

 
\subsection{Tuning and RLXF}
Continued domain-specific SFT has shown to be effective for enhancing factuality, particularly in the absence of such knowledge during pre-training.
For instance, \cite{elaraby2023halo} enhance the factual accuracy of LLMs through knowledge injection (KI). Knowledge, in the form of entity summaries or entity triplets, is incorporated through SFT by either intermediate tuning, i.e. first on knowledge and then on instruction data; or combined tuning, i.e. on the mixture of both.
While some improvements are exhibited, the method alone can be insufficient to fully mitigate factual errors.

For general-purpose LLMs, SFT is typically employed to improve the instruction-following capabilities as opposed to factual knowledge which is mostly learned in pre-training. 
However, this process may inadvertently reveal areas of knowledge not covered in the pre-training, causing the risk of behavior cloning, where a model feigns understanding and responds with hallucinations to questions it has little knowledge of~\cite{torabi2018behaviorclone}.
R-tuning~\cite{zhang2023rtuning} is proposed to address this issue with two pivotal steps: first, assessing the knowledge gap between the model's parametric knowledge and the instruction tuning data, and second, creating a refusal-aware dataset for SFT.
It enables LLMs to abstain from answering queries beyond their parametric knowledge scope.
On the other hand, BeInfo~\cite{razu2023beinfo} improve factual alignment through the form of behavioral fine-tuning. 
The creation of the behavioral tuning dataset emphasizes two goals: selectivity (choosing correct information from the knowledge source) and response adequacy (informing the user when no relevant information is available or asking for clarification). Both methods effectively control LLMs on non-parametric questions but require extra effort in dataset curation and might hinder the models' retention of parametric knowledge.

Sycophancy~\cite{sharma2023sycophancy}, another source of factuality errors, often arises from misalignments during SFT and RLHF\cite{Ouyang2022TrainingLM}. This is partially attributed to human annotators' tendency to award higher scores to responses they like rather than those that are factually accurate. \cite{wei2023sycoreduce} explore the correlation of sycophancy with model scaling and instruction tuning. They propose a synthetic-data intervention method, using various NLP tasks to teach models that truthfulness is independent of user opinions. However, one limitation is that the generalizability of their approach remains unclear for varied prompt formats and diverse user opinions.

\cite{tian2023fine} utilize direct preference optimization (DPO)~\cite{rafailov2023dpo} with the feedback of factuality score either from automatic fact-checkers or LLMs predictive confidence.
In-domain evaluation shows promising results on biographies and medical queries, but generalization performance across domains and unseen domains is under-explored.
\cite{koksal2023hallucination} propose hallucination-augmented recitations (HAR). 
It encourages the model to attribute to the contexts rather than its parametric knowledge, by tuning the model on the counterfactual dataset created leveraging LLM hallucinations.
This approach offers a novel way to enhance LLM attribution and grounding in open-book QA.
However, challenges lie in refining counterfactual generation for consistency and expanding its application to broader contexts.

\paragraph{Retrieval Augmentation} 
Incorporating retrieval mechanisms during fine-tuning has been shown to enhance the LLM factuality on downstream tasks, particularly in open-domain QA. 
DPR~\cite{Karpukhin2020DensePR} refines a dual-encoder framework, consisting of two BERT models. It employs a contrastive loss to align the hidden representations of questions and their corresponding answers, obtained through the respective encoder models.
RAG \cite{Lewis2020RetrievalAugmentedGF} and FiD \cite{Izacard2020LeveragingPR} study a fine-tuning recipe for retrieval-augmented generation models, focusing on open-domain QA tasks.
WebGPT \cite{Nakano2021WebGPTBQ} fine-tunes GPT-3 \cite{brown2023gpt3} by RLHF, providing questions with factually correct long-form reference generation. The implementation in a text-based web-browsing environment allows the model to search and navigate the web.

\subsection{Inference}
We categorize approaches to improve factuality during inference into two: (1) optimizing decoding strategies to strengthen model factuality; and (2) empowering LLM learned ability by either in-context learning (ICL) or self-reasoning.

\subsubsection{Decoding Strategy}
Sampling from the top subword candidates with a cumulative probability of p, known as nucleus sampling (top-p)~\cite{holtzman2020curious}, sees a decrease in factuality performance compared to greedy decoding, despite higher diversity. This is likely due to its over-encouragement of randomness.
Building on the hypothesis that sampling randomness may damage factuality when generating the latter part of a sentence than the beginning, \cite{lee2022factuality} introduce factual-nucleus sampling, 
which dynamically reduces the \textit{nucleus-p} value as generation progresses to limit diversity and improve factuality,
modulating factual integrity and textual diversity.

Apart from randomness, some errors arise when knowledge conflicts, where context contradicts information present in the model's prior knowledge.
Context-aware decoding (CAD)~\cite{shi2023trusting} prioritizes current context over prior knowledge, and employs contrastive ensemble logits, adjusting the weight of the probability distribution when predicting the next token with or without context.
Despite the factuality boost, CAD is a better fit for tasks involving knowledge conflicts and heavily reliant on high-quality context.

In contrast, DoLa~\cite{chuang2023dola} takes into account both upper and lower (earlier) layers, as opposed to only the final (mature) layer.
This method dynamically selects intermediate layers at each decoding step, in which an appropriate premature layer contains less factual information with maximum divergence among the subset of the early layers. 
This method effectively harnesses the distinct contributions of each layer to factual generations.
However, DoLa increases the decoding time by 1.01x to 1.08x and does not utilize external knowledge, which limits its ability to correct misinformation learned during training.

\subsubsection{ICL and Self-reasoning}
In context learning (ICL) allows an LLM to leverage and learn from demonstration examples in its context to perform a particular task without the need to update model parameters.
\cite{zheng-etal-2023-edit} present that it is possible to perform knowledge editing via ICL through facts included in demonstration examples, thereby correcting fake or outdated facts.
The objective of demonstration examples is to teach LLMs how to:
(1) identify and copy an answer;
(2) generalize using in-context facts;
(3) ignore irrelevant facts in context.

While it is rather easy for LLMs to copy answers from contexts, changing predictions of questions related to the new facts accordingly, and keeping the original predictions if the question is irrelevant to the modified facts, remains tough.




Another line of research leverages the self-reasoning capability of LLMs.
\cite{du2023debate} improve LLM factuality through multi-agent debate.
This approach first instantiates a number of agents and then makes them debate over answers returned by other agents until a consensus is reached.
One interesting finding is that more agents and longer debates tend to lead to better results. 
This approach is orthogonal and can be applied in addition to many other generation methods, such as complex prompting strategy (e.g., CoT~\cite{wei2022cot}, ReAct~\cite{yao2023react}, Reflexion~\cite{shinn2023reflexion}) and retrieval augmentation.





\textit{\textbf{Take-away:}}
\citet{zheng-etal-2023-edit} evaluate the effectiveness of knowledge editing on subject-relation-object triplets, an unrealistic setting compared to open-ended free-form text assessment. Previous methods \cite{Mitchell2021FastME, Meng2022LocatingAE} use finetuning over texts containing specific text to improve factuality. The relationship between SFT and ICL may also been an interesting avenue to explore.
More specifically,  we seek answers to two research questions:
(1) What types of facts and to what extent can facts be edited effectively, learned by LLMs through ICL? (2) Would SFT do a better job at learning from examples that are difficult for ICL? More broadly, what is the best way to insert new facts or edit false knowledge stored in LLMs. The community may also benefit from an in-depth comparative analysis of the effectiveness of improving factuality between SFT and ICL (perhaps also RLXF).


\paragraph{Retrieval Augmentation}
can be applied before, during, and after model generation.

One commonly used option is to apply retrieval augmentation prior to response generation.
For questions requiring up-to-date world knowledge to answer, \cite{vu2023freshllms} augment LLM prompts with web-retrieved information and demonstrate the effectiveness on improving accuracy on \freshqa, where \chatgpt and \gptfour struggle due to their lack of up-to-date information.
\cite{gao-etal-2023-enabling} place all relevant paragraphs in the context and encourage the model to cite supporting evidence, instructing LLMs to understand retrieved documents and generate correct citations, thereby improving reliability and factuality. 

Pre-generation retrieval augmentation is beneficial as the generation process is conditioned on the retrieval results, implicitly constraining the output space.
While improving factual accuracy, this comes at the cost of spontaneous and creative responses, largely limiting the capabilities of LLMs.
An alternative method is to verify and rectify factual errors after the model generates all content.
However, LLMs have been shown to be susceptible to hallucination snowballing~\cite{zhang2023snowball}, a common issue where a model attempts to make its response consistent with previously generated content even if it is factually incorrect.


Striking a balance between preserving creative elements and avoiding error propagation,
EVER~\cite{kang2023ever} and ``a stitch in time saves nine''~\cite{varshney2023stitch} actively detect and correct factual errors \textbf{during generation} sentence by sentence. 
The former leverages retrieved evidence for verification, and the latter incorporates the probability of dominant concepts in detection. 
Their findings suggest that timely correcting errors during generation can prevent snowballing and further improve factuality.
Nonetheless, the primary concern for this iterative process of generate-verify-correct in real-time systems is latency, making it difficult to meet the high-throughput and responsiveness demand~\cite{kang2023ever}.

\subsection{Automatic Fact Checkers}
\begin{figure}[t!]
	\centering
	\includegraphics[scale=0.3]{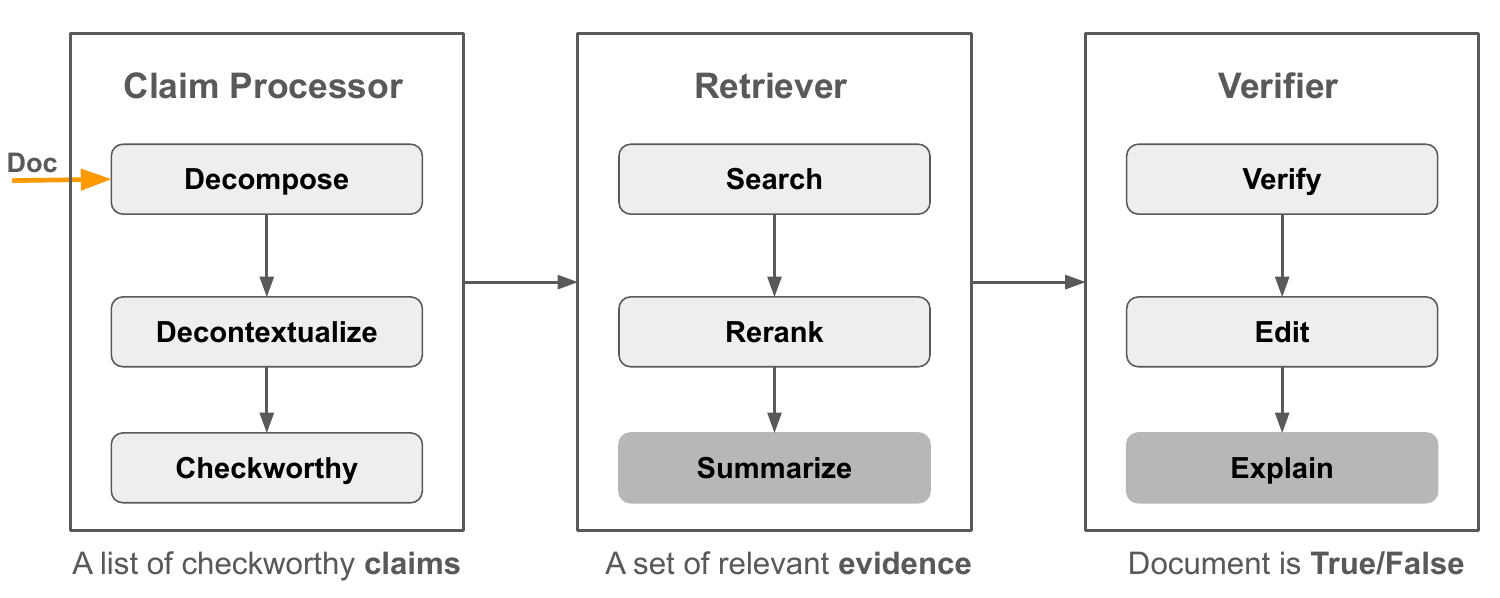}
	\caption{Fact-checker framework: claim processor, retriever, and verifier, with optional step of \textit{summarizing} and \textit{explaining} in gray.}
	\label{fig:fact-checker}
\end{figure}
An automatic fact-checking framework typically consists of three components: claim processor, retriever, and verifier as shown in \figref{fig:fact-checker}, though the implementation of verification pipelines may differ.
For example, \factor~\cite{muhlgay2023factor} and \factscore~\cite{min2023factscore} only detect falsehoods without correction. 
While \rarr depends on web-retrieved information~\cite{gao2022attributed}, and \cove~\cite{dhuliawala2023chain} only relies on LLM parametric knowledge~\cite{dhuliawala2023chain} to perform both detection and correction, albeit at a coarse granularity, editing the entire document.
Compared to fine-grained verification over claims, it is unable to spot false spans precisely and tends to result in poor preservation of the original input.
\factool~\cite{chern2023factool} and \factcheckgpt~\cite{wang2023factcheckgpt} edit atomic claims.
While the former breaks a document down to independent checkworthy claims with three steps: decomposition, decontextualization and checkworthiness identification, the latter employs \gptfour to extract verifiable claims directly.
Evaluating the effectiveness of fact-checkers remains challenging, making the improvement of such systems a difficult task.

\paragraph{Engineering and Practical Considerations}
Automatic fact-checking involve tasks of extracting atomic check-worthy claims, collecting evidence either by leveraging the knowledge stored in the model parameters or retrieved externally, and verification.
While straightforward to implement, this pipeline may be susceptible to error propagation.
Major bottleneck lies in the absence of automatic evaluation measures to assess the quality of intermediate steps, in particular, the claim processor and evidence retriever as there is no gold standard.

The input to a claim processor is a document and the expected output is a list of atomic check-worthy claims or atomic verifiable facts.
There is no consensus on the granularity of ``atomic claims'', making consistent decomposition difficult.
Additionally, the concept of check-worthy and verifiable claims are subjective.
Consequently, the definition of an atomic check-worthy claim remains a highly debatable topic.
This naturally leads to different ``gold'' human-annotated atomic claims annotated following various guidelines and distinct implementation approaches to decompose a document.

Given a document, even if assuming a ground-truth list of atomic claims, it is an open question how to assess the quality of automatically derived decomposition results.
\cite{wang2023factcheckgpt} assess the agreement in the number of claims between ground truth and predictions, followed by examining the semantic similarity between two claims at the same index when the claim count aligns.
Entailment ratio presented in \secref{sec:othermetrics} is also applicable~\cite{lee2022factuality}.

While it is much simpler when the evidence is constrained (e.g., to Wikipedia documents as is the case for FEVER~\cite{thorne-etal-2018-fever}), accurate retrieval of evidence from the Internet and subsequently quantifying the quality of such retrieval results remain challenging.
Similar to the assessment of atomic claims, gold-labeled evidence is unavailable and infeasible to obtain in the expansive open search space.


The only step where we can confidently evaluate its quality is the accuracy of verification, a simple binary true/false label given a document/claim.
In conclusion, perhaps the most significant hurdle for the development and improvement of automatic fact-checkers lies in the automated assessment and quantification of the quality at intermediate stages.

\section{Factuality of Multimodal LLMs}
\label{sec:modality}
 Factuality or hallucination in Multimodal Large Language Models refers to the phenomenon of generated responses being inconsistent with the image content. Current research on multimodal factuality can be further categorized into three types:
\begin{enumerate}[noitemsep]
    \item \textit{Existence Factuality}: incorrectly claiming the existence of certain objects in the image.
    \item \textit{Attribute Factuality}: describing the attributes of certain objects in a wrong way, e.g. identifying the colour of a car incorrectly.
    \item \textit{Relationship Factuality}: false descriptions of relationships between objects, such as relative positions and interactions.
\end{enumerate}

\paragraph{Evaluation} CHAIR~\citep{Rohrbach2018ObjectHI} is the first benchmark for assessing the accuracy of object existence within captions, focusing on a predefined set of objects in the COCO dataset~\citep{Lin2014MicrosoftCC}. However, this approach can be misleading since the COCO dataset is frequently used in training sets, providing a limited perspective when used as the sole basis for evaluation. In contrast, POPE~\cite{Li2023EvaluatingOH} evaluates object hallucination with multiple binary choice prompts, both positive and negative, querying if a specific object exists in the image. More recently, \cite{Li2023EvaluatingOH} proposed \emph{GPT4-Assisted Visual Instruction Evaluation (GAVIE)} to evaluate the visual hallucination  Additionally, \cite{Gunjal2023DetectingAP} demonstrated the use of human evaluation to avoid inaccuracies and systematic biases.

\paragraph{Mitigation} The methods for improving factuality in MLLMs can be broadly categorized into the categories: finetuning-based method, inference time correction and representation learning. 

Fine-tuning methods such as LRV-Instruction~\cite{Liu2023MitigatingHI} and LLaVA-RLHF~\cite{Sun2023AligningLM} follow an intuitive and straightforward solution of collecting specialized data such as positive and negative instructions or human preference pairs. This data is used for finetuning the model, thus resulting in models with fewer hallucinated responses.  Whereas inference time approaches mitigate factuality by correcting output generation. Woodpecker~\cite{Yin2023Woodpecker} and LURE~\cite{Zhou2023AnalyzingAM} use specialized models to rectify model generation. There are other works such as HallE-Switch~\cite{Zhai2023HallESwitchRA}, VCD~\cite{Leng2023MitigatingOH}, and HACL~\cite{Jiang2023HallucinationAC} that analyse and improve feature representation to improve factuality.
\section{Challenges and Future Directions}
\label{sec:challenges}
We first identify three major challenges for improving the factuality of LLMs, and then we point to several promising directions for future work.

\paragraph{Language models learn a language distribution, not facts.}
The training objective of language modeling is to maximize the probability of a sentence, as opposed to that of a factual statement.
While capable of generating seemingly coherent and fluent outputs upon convergence, models are not guaranteed to always return a factual response.

\paragraph{Automatic evaluation of the factual accuracy of open-ended generations remains challenging.}
Existing studies on factuality enhancement use different benchmarks and evaluation measures, making fair comparisons difficult, which motivates the need for a unified automated evaluation framework that uses the same collection of datasets and metrics.
Current approaches rely on either human evaluation or results of automated fact-checkers such as FactScore and FacTool~\cite{min2023factscore,chern2023factool}.
However, automatically quantifying the quality of automated fact-checkers is itself an open question, resulting in a chicken and egg situation.

\paragraph{Latency and multi-hop reasoning could be the bottleneck of RAG systems.}
Retrievers serve as the core component in RAG systems, and the effectiveness of RAGs is largely influenced by the quality (coverage and relevance) of the retrieved documents.
Latency and difficulties in gathering the most pertinent evidence are the primary challenges in retrieval. 
While this is partly due to the inability of ranking algorithms to retrieve such documents, certain facts require information gathered from various sources and multi-hop reasoning.

\paragraph{\textit{Potential Future Directions}}
\textit{Mitigation in inference:}
We observe that models can often generate a correct answer in multiple trials even if some attempts are wrong~\cite{tian2023fine}.
This motivates us to ask how to provide an anchor that can guide LLM decoding to the factually correct path?

Iteratively detecting, correcting, and generating during generation has been demonstrated to be effective to mitigate hallucinations.
If simply correcting the first one or two sentences, how much improvements can we expect for subsequent generations?
Can factually correct and relevant sentences, phrases or concepts serve as anchors?

\textit{Development of better retrieval algorithms:}
Integrating Retrieval-Augmented Generation (RAG) into Large Language Models (LLMs) is challenging due to the prevalence of unreliable information, such as fake news, on the internet. This compromises the accuracy of the knowledge retrieved, resulting in LLMs generating responses based on incorrect input. Consequently, future research should focus on improving retrieval techniques to enhance the factuality of LLM-generated responses.


\textit{Improving the efficiency and the accuracy of automated fact-checkers:}
The key breakthrough in effectively evaluating the factual accuracy of LLMs lies in establishing accurate and efficient fact-checkers. This requires improvement of the quality of the evidence used for making veracity decisions. Moreover, many recent methods rely on the factuality of stronger models such as GPT-4 for claim verification. Not only is this computationally expensive, but it also tends to be highly sensitive to minor prompt changes and LLM updates. A small task-specific and well fine-tuned NLI model can be a more viable, robust, and cost-efficient option.


\vspace{-1mm}
\section{Conclusion}
\vspace{-1mm}
\label{sec:conclusion}
We presented an overview on the factuality of LLMs, surveying a number of studies covering topics such as evaluation 
and improvement methods (applicable at various stages: pre-training, SFT, inference and post-processing) 
along with their respective challenges.
We also identified three major issues and pointed out to promising future research directions.
\vspace{-1mm}
\section*{Limitations}
\vspace{-1mm}
Despite conducting an extensive literature review to encompass all existing research on the factuality of LLMs, some studies may have been omitted due to the rapidly evolving nature of this research area. We endeavored to include all pertinent studies and references wherever feasible. This survey only briefly touches upon the factuality issues associated with vision language models. However, there is room for a more in-depth exploration of mitigation techniques specific to vision-language models. Additionally, comprehensive discussions are also necessary for language models that incorporate other modalities, such as video and speech.





\bibliography{custom}
\appendix

\end{document}